\documentclass[runningheads]{llncs}
\usepackage{graphicx}
\usepackage{subcaption}
\usepackage{amsmath}
\usepackage{amssymb}
\usepackage{booktabs}
\usepackage{cite}
\begin{document}
%
\title{Should Graph Neural Networks Use\\Features, Edges, Or Both?}
%
%
\author{Lukas Faber
\and Yifan Lu
\and Roger Wattenhofer
}
\authorrunning{Faber, Lu, Wattenhofer}
%
\institute{ETH Zurich, Switzerland
}
\maketitle              

\begin{abstract}
Graph Neural Networks (GNNs) are the first choice for learning algorithms on graph data. GNNs promise to integrate (i) node features as well as (ii) edge information in an end-to-end learning algorithm.
How does this promise work out practically? In this paper, we study to what extend GNNs are necessary to solve prominent graph classification problems. We find that for graph classification, a GNN is not more than the sum of its parts. We also find that, unlike features, predictions with an edge-only model do not always transfer to GNNs.
\end{abstract}

\section{Introduction}
Many complex real-world systems can be modeled as a graph, a set of nodes connected by edges.
Graph Neural Networks (GNNs) have become a method of choice to apply learning in graphs. On a high level, each node comes with an initial set of features, which a GNN then translates into an embedding. For several rounds, nodes exchange and update their embeddings with their graph neighbors.
The entire operations are differentiable, making GNNs essentially combine (i) the original \textit{features} and (ii) the \textit{edges} of the graph into an end-to-end learning algorithm. 
\begin{figure}[b]
\includegraphics[width=0.95\textwidth]{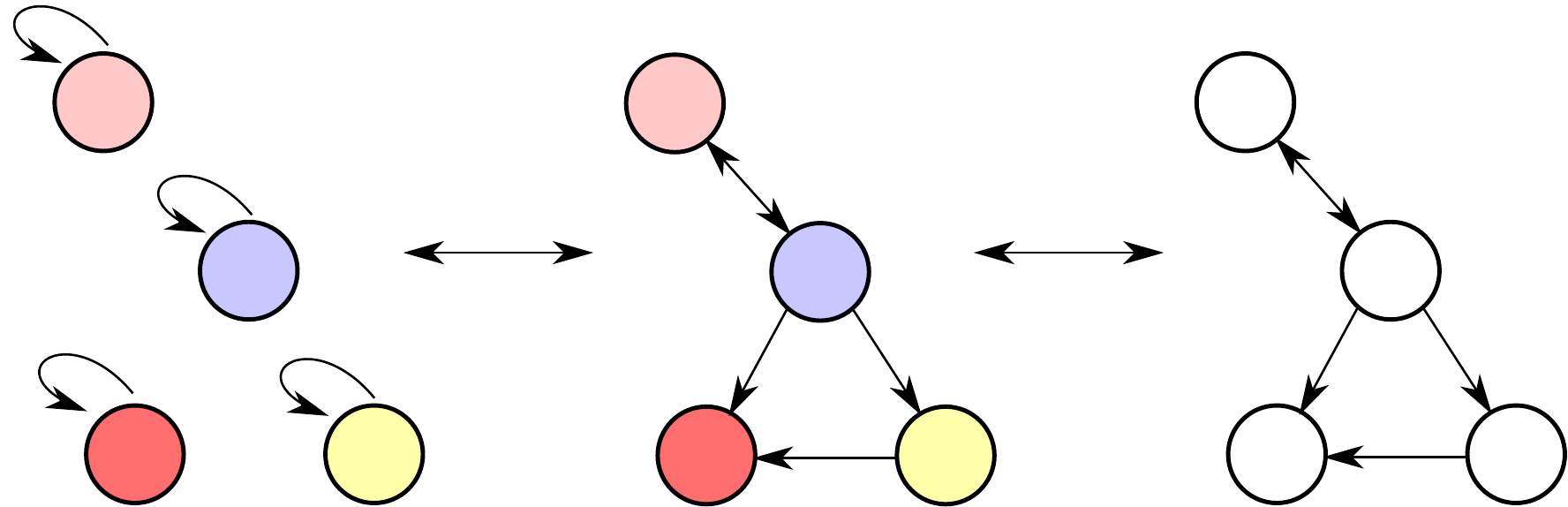}
\caption{A GNN (center) is the composition of a feature-only model (left) and an edge-only model (right).}
\label{fig_models}
\end{figure}
In this paper we study whether this intuition about GNNs is true. Do the prominent GNN problems use both (i) features \textit{and} (ii) edges? To what extent can a dataset be solved with only features \textit{or} edges? How well do GNNs to exploit the combination of features and edges, i.e., is the combination more than the sum of the parts (as in Figure~\ref{fig_models})?

The contributions of this paper can be summarized as follows:
\begin{itemize}
    \item We present two measures, FandE (\underline{F}eatures \underline{and} \underline{E}dges) and ForE (\underline{F}eatures \underline{or} \underline{E}dges) to quantify for a dataset to what extent we can use edges and features to solve its predictions. A high ForE score means that GNNs are not necessary for solving a dataset.
    \item In experiments on popular datasets, we find that several eminent datasets are largely solved by \textit{only} features or edges and thus do not need GNNs. 
    We believe that these datasets should be supplemented with datasets with a lower ForE.
    \item We further present GaP (\underline{G}NNs \underline{a}nd its \underline{P}arts). GaP measures to what extent a GNN can solve predictions neither features nor edges can solve and to what extent GNNs can also solve what either of those models could solve.
    \item In experiments on existing architectures and datasets, we show that current GNN models are better in node classification than graph classification.
\end{itemize}

\section{Related Work}
\subsection{GNN Architectures}
One of the pioneering works is the GNN model by Scarselli et al.~\cite{scarselli2008graph}. There, nodes have an embedding that is propagated and updated with their neighbors until convergence. The final embeddings are used for prediction, receive an error signal, and update. Since then, many further GNN models were proposed that generally follow the message passing paradigm~\cite{gilmer2017neural}. Similar to Scarselli et al., nodes in this framework have embeddings, they create messages based on these embeddings, send those messages to their neighbors, and update their embeddings based on received messages. There are many different instantiations of this message passing framework~\cite{wu2020comprehensive}, for example some based on spectral graph theory~\cite{kipf2017semisupervised, defferrard2016convolutional}, (self-) attention ~\cite{lee2019self, velickovic2018graph}, convolution operations~\cite{niepert2016learning, xu2019how} or residual connections~\cite{xu2018jumping}.

\subsection{Datasets for GNN Evaluation}
Currently, many GNN architectures are evaluated on the same datasets. However, current research investigates how appropriate these datasets are for benchmarking GNNs. Zou et al.~\cite{zou2019dimensional} show that the prominent Cora and Citeseer datasets leak label information in their features, thus allowing several shortcuts in predicting node labels. Huang et al.~\cite{huang2020combining} highlight another property of commonly used node classification datasets: homophily. Exploiting this property with an enhanced label propagation algorithm, they can achieve superior performance over GNNs. Zhu et al.~\cite{zhu2020beyond} examine the interplay between homophily in datasets and GNN performance. They find many current datasets to be biased towards homophily. In this paper, we propose a novel measure, ForE, to quantify the difficulty of a dataset (if we can solve the dataset without GNNs) and find that several prominent datasets are indeed easy to solve using only features \textit{or} edges. We argue future GNN evaluation should rely on more challenging datasets. There are several data collections available~\cite{hu2020open, mcauley2015imagebased, sinha2015mag, morris2020tudataset, dwivedi2020benchmarking}, from which we can further assemble datasets and evaluate those datasets with ForE.

\subsection{Limitations of current GNNs}
Despite the numerous new GNN architectures in recent years, the current theoretical and practical shortcomings in GNNs will require better architectures. From a theoretical perspective, GNNs cannot fully differentiate between different structures~\cite{xu2019how, garg2020generalization}, thus they do not understand the full capabilities of graphs. From a practical perspective, most current GNN architectures are limited to shallow models or suffer from the oversmoothing effect~\cite{li2018deeper, oono2020graph, zhao2020pairnorm, chen2020measuring}. Thus, models are limited to few layers. This restricts every node to extract local information only. Furthermore, existing empirical differences between GNN architectures are smaller than thought~\cite{shchur2018pitfalls}. To foster further evaluation of GNNs, we contribute a new dimension on how to go forward evaluating GNNs parallel to benchmark high-scores. A GNN should encompass the predictions of its parts: features and edges. This measure shows already one potential angle of improvement since current GNN architectures cannot always preserve the predictions of an edge-only model.

\section{Background}
\subsection{Examined Problems.} In this paper, we consider the two prominent tasks in graph machine learning: Node classification and graph classification. In node classification, we usually have a single graph that is partially labeled. The aim is to train a model to classify all unlabeled nodes. In graph classification, we have a collection of training graphs where each graph has a label. The aim is to train a model to learn to classify further unseen graphs. Here, we unify both tasks under the framework of having to make predictions. Thus, a prediction can be assigned a label to a node or a graph and thus a prediction can be correct or wrong. Generally, we are interested to measure how many correct predictions a certain model does.

\subsection{Perspective on GNNs.} Almost all GNN architectures~\cite{velickovic2018graph, kipf2017semisupervised, xu2019how, defferrard2016convolutional, wu2020comprehensive, hamilton2017inductive, xu2018jumping} follow the message passing framework~\cite{gilmer2017neural}. Every node starts with an embedding given by its initial features. One round of message passing then consists of nodes creating a message based on this embedding, sending this message to their neighbors, collect and aggregate all received messages, and finally updating their embedding on the received information. Usually, a GNN performs some number $k$ rounds of message passing. Thus, the node's embedding in a GNN does reflect its features and the information of its $k-$hop neighborhood. In the case of a graph classification problem, we read out a graph-level representation by summing up node embeddings. Compared to models that can only leverage either features or edges (the neighborhood information), a GNN has a superset of information. A GNN is a composition of both approaches (as in Figure~\ref{fig_models}). In this paper, we will investigate how much simpler models can sufficiently tackle various datasets and how much a GNN composition does improve over its parts.

\section{ForE and GaP}
In this section, we define the measures of ForE and GaP. Both measures are built upon investigating the predictive power of two baseline models: a feature-only model and an edge-only model, which we will define in Section~\ref{sec_models}. We measure the predictive power of these models through the \emph{set of solvable nodes (solvable set)}. We define this set in Section~\ref{sec_solvable}. Next, we introduce ForE in Section~\ref{sec_gnnn}. Here, we measure the ratio of nodes that are not in the union of the solvable sets of a feature-only and an edge-only model---that is for which we need the composed GNN model to hope for correct predictions. Last, we define GaP in Section~\ref{sec_gnnr}: The solvable set of a GNN should contain the solvable set of both a feature-only and an edge-only model, plus additional nodes. Such GNNs are indeed more than the sum of its parts.

\subsection{Feature-Only and Edge-Only Models}
\label{sec_models}
A GNN combines a feature-only model and an edge-only model into a single model. We will illustrate this concept with the example of the GCN model~\cite{kipf2017semisupervised}. Here, node embeddings for layer $l+1$ are derived according to:
\begin{equation}\label{eq_gcn_update_rule}
    H^{(l+1)} = \sigma(D^{\frac{1}{2}}AD^{\frac{1}{2}}H^{(l)}W^{(l)})
\end{equation}
For the \textbf{Feature-Only} model, we stack the node features into a $node\times feature$ input matrix and feed this matrix through a multilayer feedforward network. Thus, every set of node features gets exposed to a set of weight matrices and nonlinear activations. This is equivalent to the multiplication with $W^{(l)}$ and final activation with a nonlinearity $\sigma$ in the GCN model (see Equation~\eqref{eq_gcn_update_rule}). As alternative, less architecturally aligned choices, we could employ any ``classical'' machine learning algorithm to predict labels, given an $sample \times feature$ input matrix. One particularly interesting alternative algorithm could be a tree-based model which allows for higher interpretability and might indicate how a model can predict certain nodes.

For the \textbf{Edge-Only} model, we use the actual GNN propagation --- except we replace the initial node features with an all-ones vector. That is, in Equation~\eqref{eq_gcn_update_rule}, we replace $H^{(0)}$ with a same shaped all-ones matrix. As other edge-only models, we considered extracting features manually and providing every node these features (such as its degree, pagerank, centrality measure,\dots). However, we believe allowing the model to extract relevant features itself to be the better and more flexible approach. As another alternative, we see label propagation algorithms, as were used by Shchur et al.~\cite{shchur2018pitfalls}. However, these approaches require label information during inference time, which we cannot provide for the inductive graph classification problem.

\subsection{Solvable Sets}
\label{sec_solvable}
In this section, we define the solvable set of predictions for a model. One possible approach would be instantiating a model several times and averaging the instance accuracies. We see this approach having two drawbacks: We do not capture the consistency in which model instances predict a certain prediction correctly and second, more importantly, the score will depend on the number of classes. In the extreme case of binary classification, even a random model will predict $50\%$ correctly. Therefore, we present another model that is independent of the number of classes and identifies only consistently correct predictions.

Across many instantiations of a model (through multiple independent runs), we investigate if the model is \emph{significantly more often} correct about a prediction than what we can explain through random guessing. Therefore, per prediction, we test the null hypothesis that the different runs can only predict correctly through chance. These runs are independently initialized, therefore, we assume that the events of correct random guesses are also independent of each other. Thus we expect the random guesses to follow a Bernoulli Distribution with a success probability $p=\frac{1}{c}$ with $c$ being the number of possible classes. If we can reject the null hypothesis for a prediction, we can conclude that the data contains information that allows us to systematically predict correctly. These are the predictions we consider solved; the solvable set of a model is the set of all solved predictions. Formally, for a universe of predictions of a dataset $P$, the solvable set of a model $M$ is defined as $S(M) = \{p \in P | M \text{ solves } p\}$. Note that this measure does not directly relate to a particular instance that does the correct prediction, but having one instance (or some ensembled instances) will allow us to get the right prediction.

\subsection{ForE}
\label{sec_gnnn}
Equipped with computing the solvable sets for feature-only and edge-only models, we quantify datasets on the question: Does this dataset even need a GNN to solve? We would consider a GNN necessary for a prediction when neither the simpler feature-only or edge-only model suffices. We can quantify this as $\text{ForE}=\frac{\vert S(Features) \cap S(Edges) \vert}{\vert P \vert}$. Ideally, we evaluate GNNs on challenging datasets where all predictions would require joint feature and edge information. A more realistic stance is that we do not want all predictions solved by looking only at the features or edges. Naturally, we can address these predictions with a GNN as well, but such a dataset does not allow us to measure how much new predictive power a new GNN architecture offers. This is because the GNN just has to learn to use either the features only or the nodes. Therefore, we would like datasets to have a low ForE value.

\subsection{GaP}
\label{sec_gnnr}
As a second application for solvable sets, we now continue to answer: Are current GNNs more than the sum of their parts? A GNN has access to all the information of its parts, plus it can additionally exploit having both features at the same time. Therefore, what we would like to see are three effects: The solvable set of the GNN is a superset of the solvable set of both the feature-only and the structure-only model. We measure this property by computing the ratio $\frac{\vert S(\text{GNN})\cap S(\text{Features}) \vert}{\vert S(\text{Features}) \vert}$ and $\frac{\vert S(\text{GNN})\cap S(\text{Edges}) \vert}{\vert S(\text{Edges}) \vert}$, respectively. An effective GNN will achieve values close to $1$. Moreover, an effective GNN should exploit having the joint information of features and edges to make predictions, which neither individual model is capable of. We measure this extent as $\frac{\vert S(\text{GNN}) \cap \text{U} \vert}{\vert \text{U} \vert}$ where $U$ is the set of ``unsolved'' nodes $U = P \setminus (S(\text{Features}) \cup S(\text{Edges}))$. In an ideal world, a GNN would be able to solve all remaining predictions thus yielding a perfect score of $1$; however, we expect real values to turn out lower due to data errors, labeling noise, and similar effects.

\section{Experiments}
We investigate prominent GNN architectures and GNN benchmarking datasets using ForE and GaP. We choose to run experiments with four archetypes of established GNN models: Graph Convolutional Networks (GCN)~\cite{kipf2017semisupervised}, GraphSage (GS; pooling and mean variants)~\cite{hamilton2017inductive}, Graph Attention Networks  (GST)~\cite{velickovic2018graph}, and Graph Isomorphism Networks (GIN; max, mean, and sum variants)~\cite{xu2019how}. For datasets, we choose from a variety of dataset collections, covering a variety of domains. We use transductive node classification datasets stemming from citation networks (Cora, Citeseer, Pubmed, OGBN-Arxiv), co-purchasing graphs (AMZN-Photo, AMZN-Comp) from Amazon~\cite{mcauley2015imagebased}, and coauthorship networks based on the Microsoft Academic Graph~\cite{sinha2015mag} (MAG-Physics, MAG-CS). For choosing the subsets of the Microsoft Academic Graph and the amazon co-purchasing data, we follow the choice of Shchur et al~\cite{shchur2018pitfalls}. On the other hand, we experiment with several inductive graph classification tasks from the dataset collection of Morris et al.~\cite{morris2020tudataset}. In Mutag, the task is to predict the chemical properties of molecules. The datasets Proteins and Enzymes come from the biology domain and classify if a protein is an enzyme or which kind of enzyme, respectively. The dataset Reddit-B is about classifying Reddit threads where nodes are users and edges for answering users whether this thread is a discussion or a Q\&A session. The last dataset IMDB-M is a dataset to detect the genre of movies based on a network of actor collaborations. Most datasets come with a suggested split for the predictions into the training, validation, and test set, which we keep. In case there is no such split, we create one fixed split.

\subsection{Hyperparameter settings}
We use the implementation provided by the DGL library~\cite{wang2019dgl} for GNN implementation, keeping the hyperparameters to their default values. We set the number of layers of the GNNs to $3$ and the embedding width to be twice the number of input or output values --- whichever is larger. Several datasets have a massive number of features (for example ), which is why we upper bound the embedding size to $128$. We train the GNNs for up to $10.000$ epochs with early stopping on the validation metrics with patience of $25$. In practice, we find that training generally needs much fewer epochs, and models converge after several hundred epochs. Models are trained with the Adam optimizer~\cite{kingma2014adam} and its default parameters in the PyTorch library~\cite{paszke2015pytorch}. To ensure structural similarity with the GNNs, we employ a $3$ layer feedforward network, which we train in the same fashion. For the edge-only model, we try all GNN propagation and pick the best one per dataset. We run each dataset model configuration $100$ times with different random seeds and do the statistical test with a $99.9\%$ confidence level ($p=0.001$).

\subsection{ForE Results}
Table~\ref{tab_gnnn} shows results, for solvable sets of feature-only, edge-only models, their expected FandE (assuming independence), the actual FandE, ForE, and the solvable set of the best-performing GNN. All node classification datasets behave very similarly: 1) Using only features, we can solve a large portion of the dataset; 2) the edge-only model is a useful and mostly independent additional source of information; 3) GNNs perform better than either. The most important observation is, however, that ForE is rather high for most datasets, especially for the MAG based datasets.

For graph classification datasets, we observe that the performance for features and edges is similar. Furthermore, the solvable sets of features and edges are not independent but correlate. This suggests that there are ``easy'' graphs that we can predict either way. For almost all datasets, no GNN architecture offers an of-the-shelf-advantage over features or nodes.

\begin{table}[!ht]
\centering
\begin{tabular}{lcccccc}
    \toprule
    Dataset  & Features & Edges & $\mathbb{E}(\text{FandE})$ & FandE & ForE & GNN\\
    \midrule
        Cora & 0.586 & 0.346 & 0.203 & 0.192 & 0.74 & 0.828\\
    \midrule
    Citeseer & 0.544 & 0.412 & 0.224 & 0.235 & 0.721 & 0.699\\
    \midrule
    Pubmed & 0.693 & 0.407 & 0.282 & 0.246 & 0.854 & 0.779\\
    \midrule
    AMZN-Photo & 0.777 & 0.286 & 0.222 & 0.172 & 0.891 & 0.909\\
    \midrule
    AMZN-Comp & 0.652 & 0.391 & 0.255 & 0.235 & 0.808 & 0.809\\
    \midrule
    MAG-Physics & 0.915 & 0.507 & 0.464 & 0.475 & 0.947 & 0.949\\
    \midrule
    MAG-CS & 0.924 & 0.136 & 0.126 & 0.129 & 0.932 & 0.933\\
    \midrule
    OGBN-Arxiv & 0.658 & 0.411 & 0.271 & 0.281 & 0.788 & 0.726\\
    \midrule
    \midrule
    Mutag & 0.45 & 0.55 & 0.248 & 0.45 & 0.55 & 0.55\\
    \midrule
    Enzymes & 0.4 & 0.333 & 0.133 & 0.2 & 0.533 & 0.65\\
    \midrule
    Proteins & 0.607 & 0.643 & 0.39 & 0.607 & 0.643 & 0.616\\
    \midrule
    IMDB-M & 0.26 & 0.293 & 0.076 & 0.24 & 0.313 & 0.287\\
    \midrule
    Reddit-B & 0.76 & 0.775 & 0.589 & 0.76 & 0.775 & 0.77\\
    \bottomrule
\end{tabular}
\caption{We analyzed 8 node prediction and 5 graph prediction datasets. We report their prediction score based on Features only and Edges only in Columns 2 and 3. $\mathbb{E}(\text{FandE})$ shows the expected ratio that Features and Edges are correct, assuming independence. FandE reports the actual overlap between Features and Edges. ForE reports the ratio of predictions that at least one of Features and Edges solve. GNN shows the correct predictions for the best-performing GNN on this dataset. Let us look at the example dataset MAG-CS. From columns 2 and 3 we learn that using just features or edges we can solve $92.4\%$ or $13.6\%$ of the predictions of this dataset, respectively. Columns 4 and 5 show that the expectation of FandE is close to the actual value, suggesting that features and edges are independent sources of information. The ForE shows that $93.2\%$ of all predictions can be solved with features or edges and do not need a GNN. Last, the best-performing GNN model solves $93.3\%$ of all nodes.}
\label{tab_gnnn}
\end{table}

\subsection{GaP}
\begin{figure}
    \begin{subfigure}{0.5\textwidth}
        \centering
        \includegraphics[width=\textwidth]{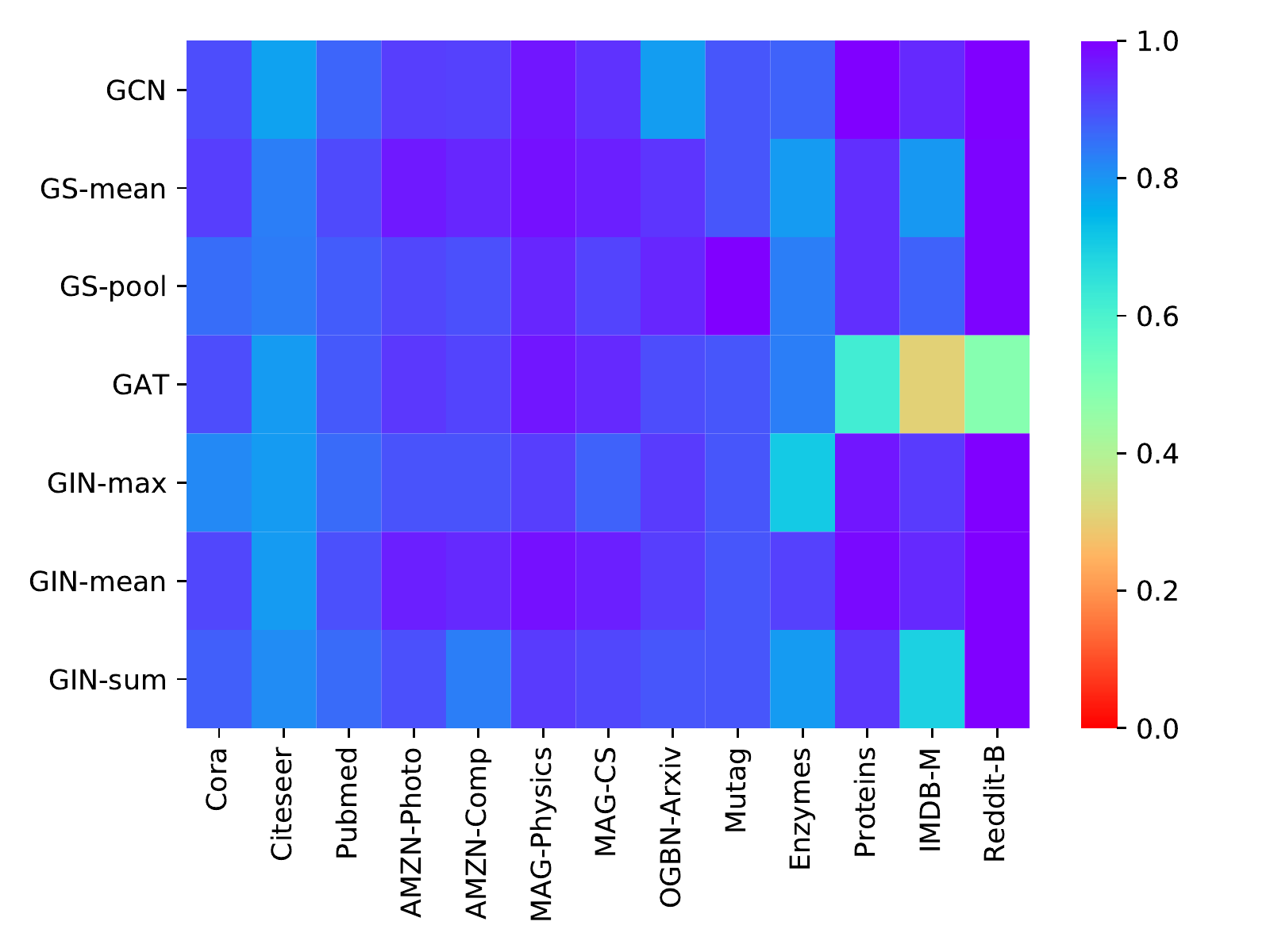}
        \caption{}
        \label{fig_featureretention}
    \end{subfigure}%
    \begin{subfigure}{0.5\textwidth}
        \centering
        \includegraphics[width=\textwidth]{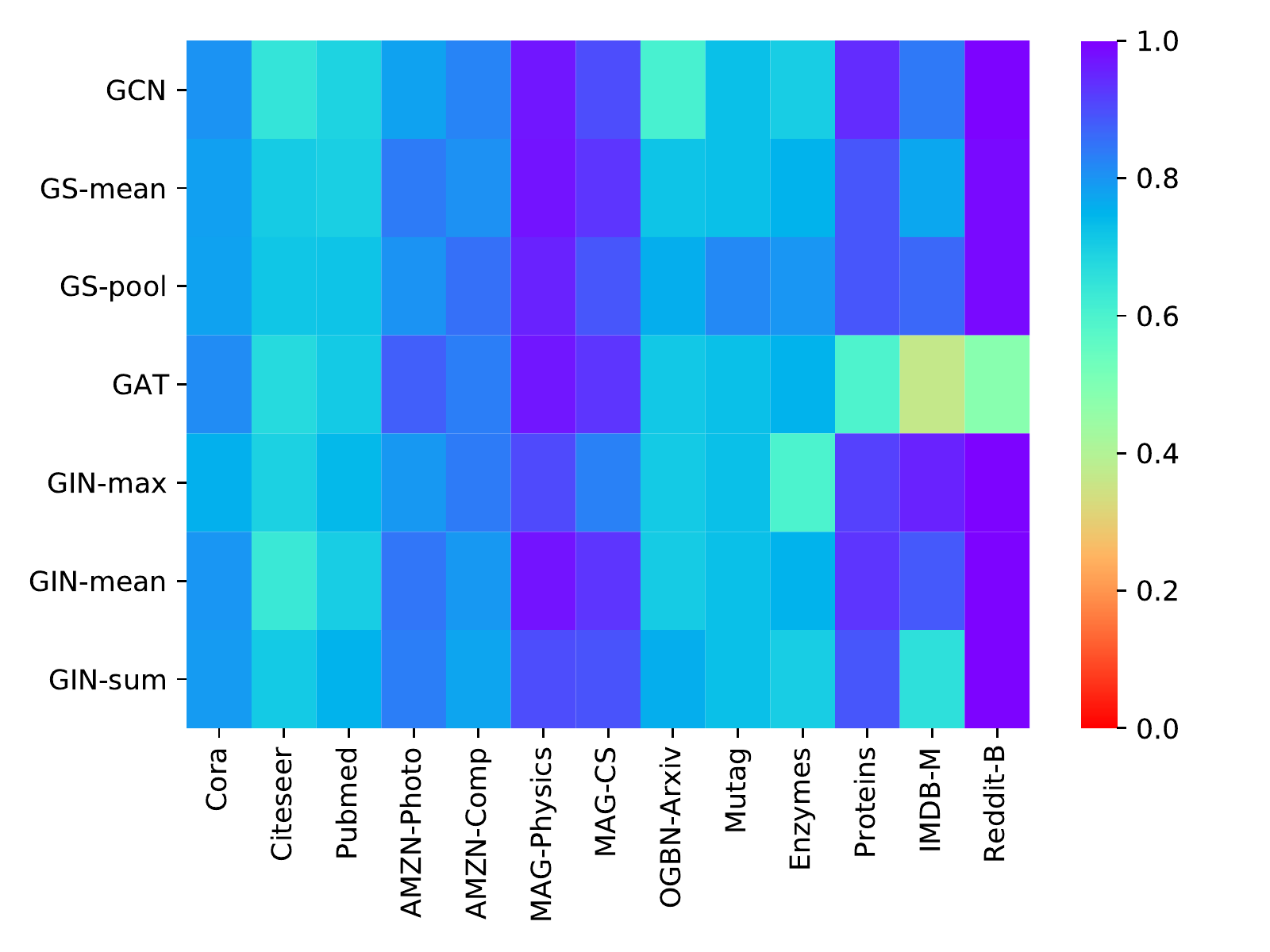}
        \caption{}
        \label{fig_edgeretention}
    \end{subfigure}
    \caption{How much can a GNN replace its parts? Per cell, we show the ratio of predictions and feature-only model (a) or an edge-only model (b) can solve that a GNN can also solve (higher values are better). For example, consider the entry for GCN and Citeseer in Figure~\ref{fig_edgeretention} (first row, second column). The score of $64.6\%$ indicates that only around $2$ out of $3$ predictions of the edge-only model solves are also solved by the GNN. This motivates new GNN architectures that preserve more predictions or cleverly ensemble GNNs with its edge-only parts.}
    \label{fig_retention}
\end{figure}

We now further investigate the performance of GNNs compared to their parts. First, we take a look if GNNs can solve those predictions its parts also solve. Figure~\ref{fig_retention} shows results for these experiments. In Figure~\ref{fig_featureretention} we can see the ratio of solved predictions the GNN preserves from the feature-only model; Figure~\ref{fig_edgeretention} shows the prediction from the edge-only model. We can observe that GNNs are consistently good at preserving feature predictions. For edge predictions the picture is less clear, for some datasets, the GNN can only preserve around two thirds. This gap allows for improving scores, for example through ensembling GNNs and feature-only models.

\begin{figure}
\begin{minipage}{0.48\textwidth}
    \centering
    \includegraphics[width=\textwidth]{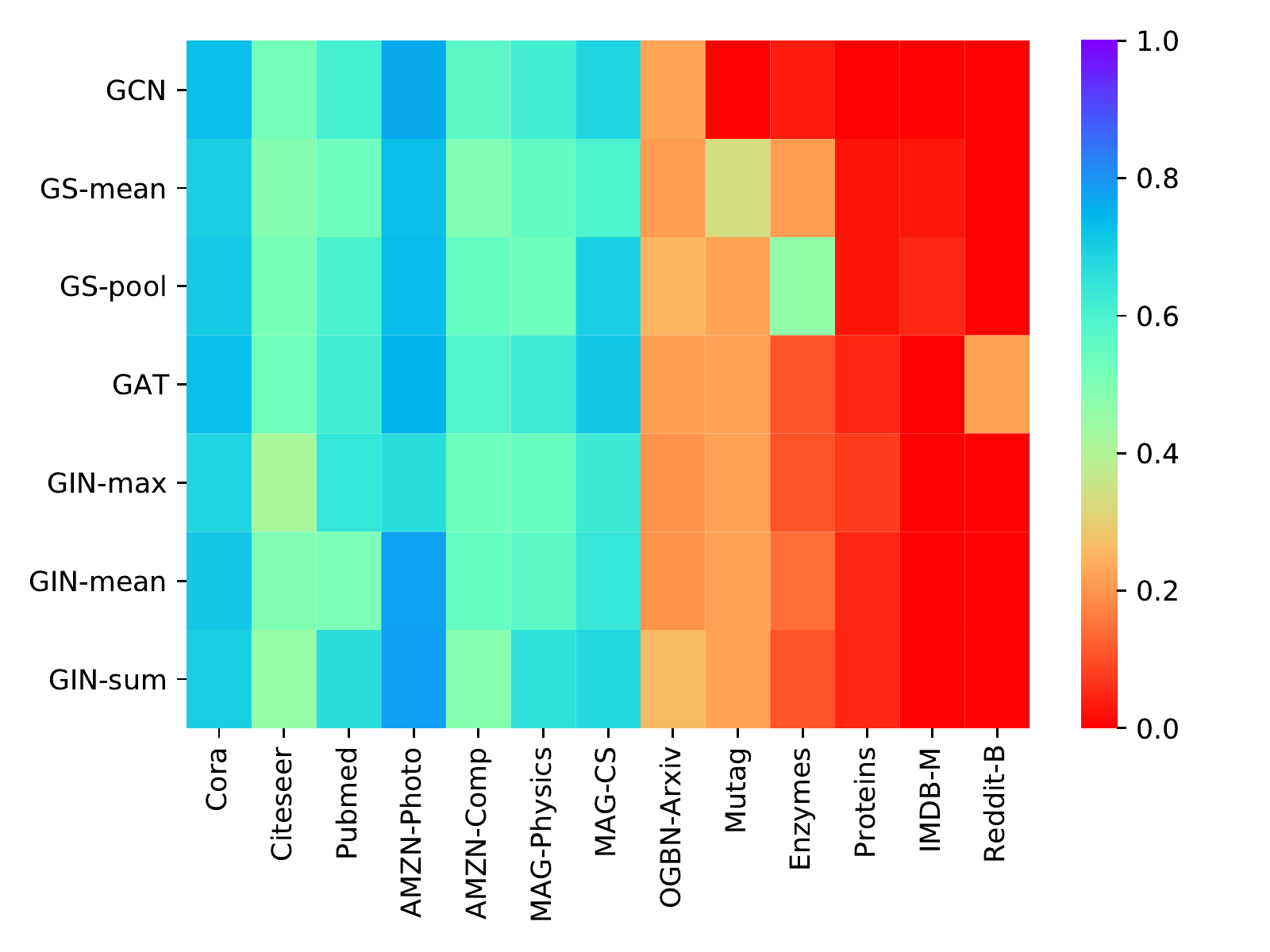}
    \caption{To what extent is a GNN more than the sum of its parts? Per cell, we show the ratio of predictions that neither features nor edges can solve, but a GNN can (higher values are better). For example, the value for GCN in Cora is $70.4\%$. This means that a GCN can solve this percentage of predictions that are not in ForE (see Table!\ref{tab_gnnn}). Thus, the GCN contributes predictions we could not do without it.}
    \label{fig_gnnadd}
\end{minipage}\hfill
\begin{minipage}{0.48\textwidth}
    \centering
    \includegraphics[width=\textwidth]{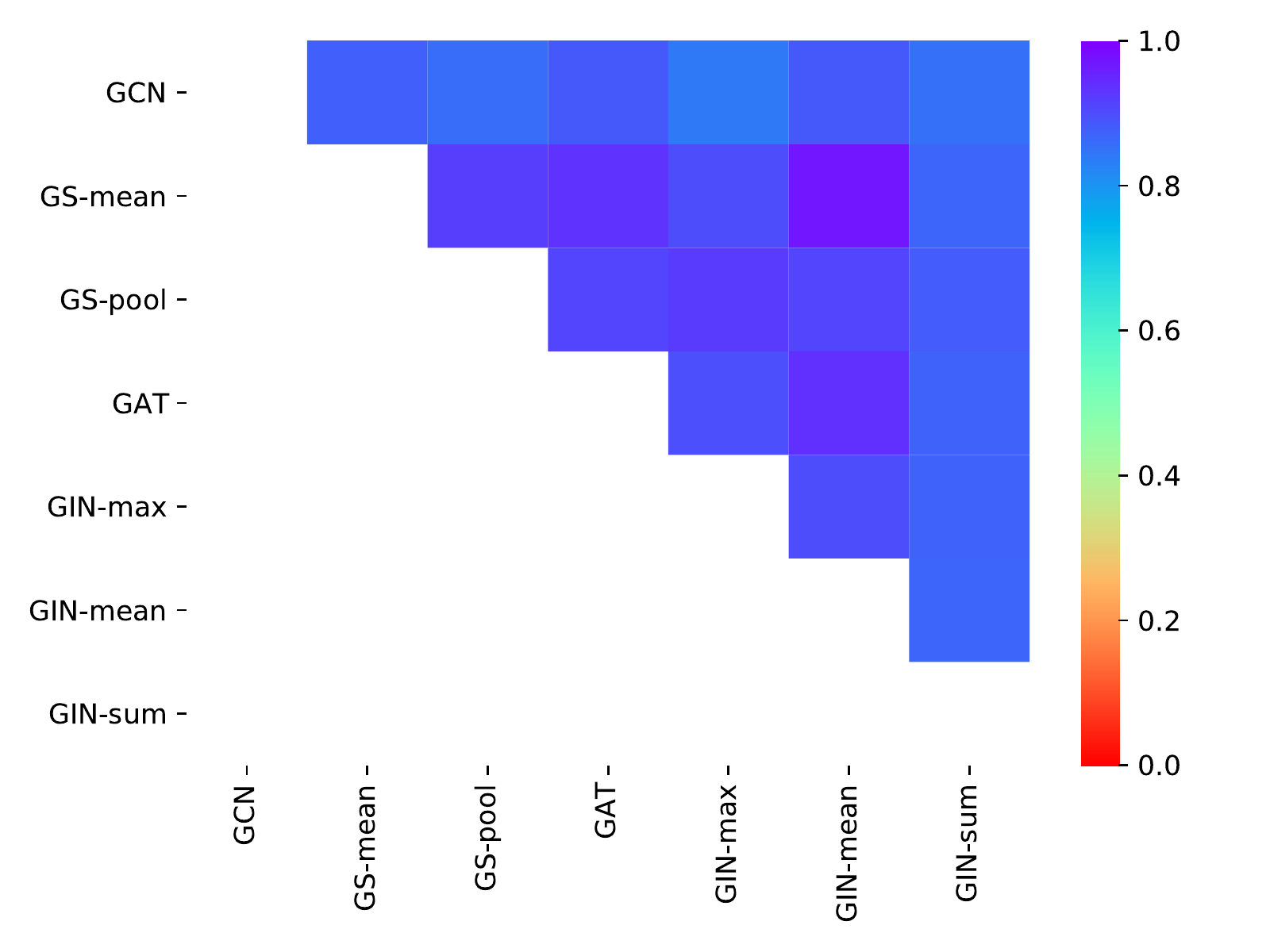}
    \caption{To what extent do solvable sets of different GNN architectures overlap? Per cell, we show the Jaccard similarity between GNN solvable sets over all datasets. For example, the Jaccard Similarity between GS-mean and GIN-mean is $97.1\%$. This shows that across all datasets, the two architectures can solve virtually the same nodes. Despite their differences in Architecture, their predictive power seems to be equal.}
    \label{fig_gnnagree}
\end{minipage}
\end{figure}

Last, we wonder if GNNs are more than the combination of their parts. We look at the ratio of the unsolved nodes that a GNN can solve in Figure~\ref{fig_gnnadd}. In all node classification datasets apart from the dataset OGBN-Arxiv, GNNs solve a large fraction of nodes neither features nor edges can do. On the other hand, we see there is a gap in expressive power for graph classification problems. GNNs do hardly solve further predictions. Across all experiments, all investigated GNN architectures perform virtually the same, except for few outliers from GAT. In general, we establish this trend in Figure~\ref{fig_gnnagree}. Every cell shows the Jaccard similarity of solvable sets across all datasets between two GNN architectures. We can see that every pair of architectures achieve similarities close to $1$. This suggests that all GNN architectures solve more or less the same predictions. This is in line with the finding of Shchur et al~\cite{shchur2018pitfalls} that find similar performance across architectures.

\section{Conclusion}
GNNs as a machine learning model provide us with an algorithm to jointly use the node features and edges. We present ForE to measure if datasets even require this capability and to what extent they can only be solved through one input. We presented GaP, to measure if GNNs can preserve the predictions of its parts and if GNNs are more than their sum. From the experiments, we draw the following conclusions:
\begin{itemize}
    \item We should rethink which datasets to use for GNN evaluation. From Zou et al.~\cite{zou2019dimensional} and Huang et al.~\cite{huang2020combining}, we know that Cora and Citeseer are not ideal datasets. Additionally, we now know that also many other node classification datasets are ``easy'' due to having a high ForE score. Especially the MAG-based datasets are virtually solved through features alone. Such datasets let us hardly measure the new expressive power of GNNs but rather a GNN's ability to harness these features.
    \item We should be able to improve current GNN performance by ensembling GNN models with edge-only models. Figure~\ref{fig_edgeretention} shows that GNNs do not preserve all correct predictions (unlike feature predictions). This also motivates further research into GNN architectures that preserve edge predictions better.
    \item Currently, GNNs are better at node classification than graph classification. We come to this conclusion because, on the one hand, GNNs outperform both feature-only and edge-only in node classification datasets (see Table~\ref{tab_gnnn}). On the other hand, GNNs generally solve more predictions than their parts for node classification (see Figure~\ref{fig_gnnadd}). This suggests that the current approach of node-level propagation and doing a graph-level readout through summation is not as effective. This motivates further research on dedicated graph classification architectures. Two example ideas are master nodes that gather global information~\cite{gilmer2017neural} or graph pooling to aggregate node embeddings~\cite{lee2019self, gao2019graph, ying2018hierarchical, cangea2018towards}.
    \item All previous observations hold almost independent of the chosen GNN architecture (the rows in Figures~\ref{fig_featureretention},~\ref{fig_edgeretention}, and~\ref{fig_gnnadd} are similar, while there are large differences per column). Furthermore, all GNN architectures yield similar observable sets (see Figure~\ref{fig_gnnagree}). Going forward, we consider it promising to again look beyond message passing to potentially diversify the solvable sets.
\end{itemize}

\bibliographystyle{splncs04}
\bibliography{main}

\begin{thebibliography}{10}
\providecommand{\url}[1]{\texttt{#1}}
\providecommand{\urlprefix}{URL }
\providecommand{\doi}[1]{https://doi.org/#1}

\bibitem{cangea2018towards}
Cangea, C., Veli{\v{c}}kovi{\'c}, P., Jovanovi{\'c}, N., Kipf, T., Li{\`o}, P.:
  Towards sparse hierarchical graph classifiers. arXiv:1811.01287  (2018)

\bibitem{chen2020measuring}
Chen, D., Lin, Y., Li, W., Li, P., Zhou, J., Sun, X.: Measuring and relieving
  the over-smoothing problem for graph neural networks from the topological
  view. In: Conference on Artificial Intelligence (AAAI), New York, USA (Feb
  2020)

\bibitem{defferrard2016convolutional}
Defferrard, M., Bresson, X., Vandergheynst, P.: Convolutional neural networks
  on graphs with fast localized spectral filtering. In: Advances in neural
  information processing systems (2016)

\bibitem{dwivedi2020benchmarking}
Dwivedi, V.P., Joshi, C.K., Laurent, T., Bengio, Y., Bresson, X.: Benchmarking
  graph neural networks. arXiv:2003.00982  (2020)

\bibitem{gao2019graph}
Gao, H., Ji, S.: Graph u-nets. arXiv:1905.05178  (2019)

\bibitem{garg2020generalization}
Garg, V., Jegelka, S., Jaakkola, T.: Generalization and representational limits
  of graph neural networks. In: International Conference on Machine Learning
  (ICML), virtual (Jul 2020)

\bibitem{gilmer2017neural}
Gilmer, J., Schoenholz, S.S., Riley, P.F., Vinyals, O., Dahl, G.E.: Neural
  message passing for quantum chemistry. In: International Conference on
  Machine Learning (ICML), Sydney, Australia (Aug 2017)

\bibitem{hamilton2017inductive}
Hamilton, W., Ying, Z., Leskovec, J.: Inductive representation learning on
  large graphs. In: Advances in neural information processing systems (2017)

\bibitem{hu2020open}
Hu, W., Fey, M., Zitnik, M., Dong, Y., Ren, H., Liu, B., Catasta, M., Leskovec,
  J.: Open graph benchmark: Datasets for machine learning on graphs.
  arXiv:2005.00687  (2020)

\bibitem{huang2020combining}
Huang, Q., He, H., Singh, A., Lim, S.N., Benson, A.R.: Combining label
  propagation and simple models out-performs graph neural networks.
  arXiv:2010.13993  (2020)

\bibitem{kingma2014adam}
Kingma, D.P., Ba, J.: Adam: {A} method for stochastic optimization. In: Bengio,
  Y., LeCun, Y. (eds.) International Conference on Learning Representations
  {ICLR} 2015, San Diego,USA (May 2015)

\bibitem{kipf2017semisupervised}
Kipf, T.N., Welling, M.: Semi-supervised classification with graph
  convolutional networks. In: International Conference on Learning
  Representations {ICLR}, Toulon, France (Apr 2017)

\bibitem{lee2019self}
Lee, J., Lee, I., Kang, J.: Self-attention graph pooling. In: International
  Conference on Machine Learning (ICML), Long Beach, USA (Jun 2019)

\bibitem{li2018deeper}
Li, Q., Han, Z., Wu, X.M.: Deeper insights into graph convolutional networks
  for semi-supervised learning. In: Conference on Artificial Intelligence
  (AAAI), New York, USA (Feb 2018)

\bibitem{mcauley2015imagebased}
McAuley, J., Targett, C., Shi, Q., van~den Hengel, A.: Image-based
  recommendations on styles and substitutes. In: ACM SIGIR Conference on
  Research and Development in Information Retrieval (SIGIR), Santiago, Chile
  (Aug 2015)

\bibitem{morris2020tudataset}
Morris, C., Kriege, N.M., Bause, F., Kersting, K., Mutzel, P., Neumann, M.:
  Tudataset: A collection of benchmark datasets for learning with graphs.
  arXiv:2007.08663  (2020)

\bibitem{niepert2016learning}
Niepert, M., Ahmed, M., Kutzkov, K.: Learning convolutional neural networks for
  graphs. In: International Conference on Machine Learning (ICML), New York,
  USA (Jun 2016)

\bibitem{oono2020graph}
Oono, K., Suzuki, T.: Graph neural networks exponentially lose expressive power
  for node classification. In: 8th International Conference on Learning
  Representations (ICLR), Addis Ababa, Ethiopia (Apr 2020)

\bibitem{paszke2015pytorch}
Paszke, A., Gross, S., Massa, F., Lerer, A., Bradbury, J., Chanan, G., Killeen,
  T., Lin, Z., Gimelshein, N., Antiga, L., Desmaison, A., Kopf, A., Yang, E.,
  DeVito, Z., Raison, M., Tejani, A., Chilamkurthy, S., Steiner, B., Fang, L.,
  Bai, J., Chintala, S.: Pytorch: An imperative style, high-performance deep
  learning library. In: Advances in Neural Information Processing Systems
  (2019)

\bibitem{scarselli2008graph}
Scarselli, F., Gori, M., Tsoi, A.C., Hagenbuchner, M., Monfardini, G.: The
  graph neural network model. IEEE Transactions on Neural Networks  (2008)

\bibitem{shchur2018pitfalls}
Shchur, O., Mumme, M., Bojchevski, A., G{\"u}nnemann, S.: Pitfalls of graph
  neural network evaluation. arXiv:1811.05868  (2018)

\bibitem{sinha2015mag}
Sinha, A., Shen, Z., Song, Y., Ma, H., Eide, D., Wang, K.: An overview of
  microsoft academic service (mas) and applications. In: International
  Conference on World Wide Web (WWW), Florence, Italy (May 2015)

\bibitem{velickovic2018graph}
Velickovic, P., Cucurull, G., Casanova, A., Romero, A., Li{\`{o}}, P., Bengio,
  Y.: Graph attention networks. In: International Conference on Learning
  Representations (ICLR), Vancouver, BC, Canada (May 2018)

\bibitem{wang2019dgl}
Wang, M., Zheng, D., Ye, Z., Gan, Q., Li, M., Song, X., Zhou, J., Ma, C., Yu,
  L., Gai, Y., Xiao, T., He, T., Karypis, G., Li, J., Zhang, Z.: Deep graph
  library: A graph-centric, highly-performant package for graph neural
  networks. arXiv:1909.01315  (2019)

\bibitem{wu2020comprehensive}
Wu, Z., Pan, S., Chen, F., Long, G., Zhang, C., Philip, S.Y.: A comprehensive
  survey on graph neural networks. IEEE Transactions on Neural Networks and
  Learning Systems  (2020)

\bibitem{xu2019how}
Xu, K., Hu, W., Leskovec, J., Jegelka, S.: How powerful are graph neural
  networks? In: International Conference on Learning Representations, {ICLR}
  2019, New Orleans, USA (May 2019)

\bibitem{xu2018jumping}
Xu, K., Li, C., Tian, Y., Sonobe, T., Kawarabayashi, K.i., Jegelka, S.:
  Representation learning on graphs with jumping knowledge networks. In:
  International Conference on Machine Learning (ICML), Stockholm, Sweden (Jul
  2018)

\bibitem{ying2018hierarchical}
Ying, Z., You, J., Morris, C., Ren, X., Hamilton, W., Leskovec, J.:
  Hierarchical graph representation learning with differentiable pooling. In:
  Advances in neural information processing systems (2018)

\bibitem{zhao2020pairnorm}
Zhao, L., Akoglu, L.: Pairnorm: Tackling oversmoothing in gnns. In:
  International Conference on Learning Representations (ICLR) 2020, Addis
  Ababa, Ethiopia (Apr 2020)

\bibitem{zhu2020beyond}
Zhu, J., Yan, Y., Zhao, L., Heimann, M., Akoglu, L., Koutra, D.: Beyond
  homophily in graph neural networks: Current limitations and effective
  designs. Advances in Neural Information Processing Systems  (2020)

\bibitem{zou2019dimensional}
Zou, X., Jia, Q., Zhang, J., Zhou, C., Yang, H., Tang, J.: Dimensional
  reweighting graph convolutional networks. arXiv:1907.02237  (2019)

\end{thebibliography}

\end{document}